\documentclass{article}

\usepackage{microtype}
\usepackage{graphicx}
\usepackage{subfigure}
\usepackage{booktabs} %

\usepackage{hyperref}

\usepackage[accepted]{icml2025}

\usepackage{amsmath}
\usepackage{amssymb}
\usepackage{mathtools}
\usepackage{amsthm}

\usepackage[capitalize,noabbrev]{cleveref}

\theoremstyle{plain}

\theoremstyle{definition}

\theoremstyle{remark}

\usepackage[textsize=tiny]{todonotes}

\usepackage{booktabs}
\usepackage{siunitx}
\usepackage{etoolbox}
\renewrobustcmd{\bfseries}{\fontseries{b}\selectfont}
\renewrobustcmd{\boldmath}{}
\usepackage{multirow}
\usepackage{xspace}
\usepackage{amssymb}
\usepackage{pifont}
\usepackage[normalem]{ulem}
\usepackage{caption}

\newcommand{\cifairX}{ciFAIR-10\xspace}
\newcommand{\cifairC}{ciFAIR-100\xspace}
\newcommand{\stl}{STL-10\xspace}
\newcommand{\imagenet}{ImageNet\xspace}
\newcommand{\hydramix}{HydraMix\xspace}
\newcommand{\methodname}{HydraMix\xspace}
\newcommand{\chimeramix}{ChimeraMix\xspace}
\newcommand{\chimeramixgrid}{ChimeraMix+Grid\xspace}
\newcommand{\chimeramixseg}{ChimeraMix+Seg\xspace}

\newcommand{\tick}{\ding{52}}

\newcommand{\cross}{--}
\newcommand\norm[1]{\left\lVert#1\right\rVert}

\newcommand{\ts}[1]{\mathbf{#1}}

\newcommand{\std}[1]{{\fontsize{6}{7.2}\selectfont ±#1}} %

\icmltitlerunning{HydraMix: Multi-Image Feature Mixing for Small Data Image Classification}

\begin{document}

\twocolumn[
\icmltitle{HydraMix: Multi-Image Feature Mixing for Small Data Image Classification}

\icmlsetsymbol{equal}{*}

\begin{icmlauthorlist}
    \textbf{
        Christoph Reinders \quad Frederik Schubert \quad Bodo Rosenhahn\\
Institute for Information Processing / L3S, Leibniz University Hannover
    }
\end{icmlauthorlist}

\icmlkeywords{Small Data, Image Classification, Image Generation, Generative Adversarial Network}

\vskip 0.3in
]

\begin{abstract}
    Training deep neural networks requires datasets with a large number of annotated examples.
    The collection and annotation of these datasets is not only extremely expensive but also faces legal and privacy problems. These factors are a significant limitation for many real-world applications.
    To address this, we introduce \methodname, a novel architecture that generates new image compositions by mixing multiple different images from the same class.
    HydraMix learns the fusion of the content of various images guided by a segmentation-based mixing mask in feature space and is optimized via a combination of unsupervised and adversarial training.
    Our data augmentation scheme allows the creation of models trained from scratch on very small datasets.
    We conduct extensive experiments on ciFAIR-10, STL-10, and ciFAIR-100.
    Additionally, we introduce a novel text-image metric to assess the generality of the augmented datasets.
    Our results show that \methodname outperforms existing state-of-the-art methods for image classification on small datasets.
\end{abstract}

\section{Introduction}

Pretraining on large datasets has led to state-of-the-art results in computer vision and machine learning \citep{krizhevskyImageNetClassificationDeep2012}.
However, the size of these datasets makes them difficult to maintain, i.e., their labeling process is costly and difficult for new applications.
This results in several problems regarding the privacy of individuals \citep{yangFairerDatasetsFiltering2020}, unclear copyright claims \citep{hendersonFoundationModelsFair2023}, and the resources required to use them \citep{bommasaniOpportunitiesRisksFoundation2021}.
With transfer learning \citep{neyshaburWhatBeingTransferred2020}, i.e., finetuning of models that have been pretrained on such a large dataset, it is possible to circumvent the resource issue.
Nevertheless, this approach leaves the legal concerns open and also makes the users of such pretrained models dependent on companies or institutions that can afford to train them \citep{sharirCostTrainingNLP2020}.
Thus, the research on making deep learning approaches work using only a small amount of data is vital.

One solution to achieving good performance with little data lies in combinatorics.
Even with a handful of examples, it is possible to combine them in effectively infinitely different ways to enable the successful training of current machine learning pipelines with the augmented dataset.
In image classification, several ways exist to mix multiple images using approaches ranging from simple image manipulations \citep{zhangMixupEmpiricalRisk2018} to methods that apply saliency guidance \citep{daboueiSuperMixSupervisingMixing2021} to create combinations that resemble real objects more closely.
However, simple methods are not able to create sensible combinations of the given objects, i.e., combinations that resemble real objects from the class or some sub-class.
Moreover, the more complex methods fail in the small data regime due to the limited number of samples.

In this work, we present \methodname, a method that is able to create meaningful combinations of images given only a handful of data samples without any pretraining.
This work extends our conference article on the ChimeraMix method \citep{reindersChimeraMixImageClassification2022}, which learns to combine the features of two images guided by a mask before decoding them into a new image.
The contributions of the original publication are:
\begin{itemize}
    \item We propose a generative approach for addressing the small data image classification task.
    \item Our generator introduces a feature-mixing architecture for combining two images. Guided by a mask, the generator learns to combine image pairs. %
    \item Experiments on benchmark datasets demonstrate that ChimeraMix outperforms previous methods in small data image classification.
\end{itemize}

\begin{figure*}[t]
    \centering
    \includegraphics[width=0.99\linewidth]{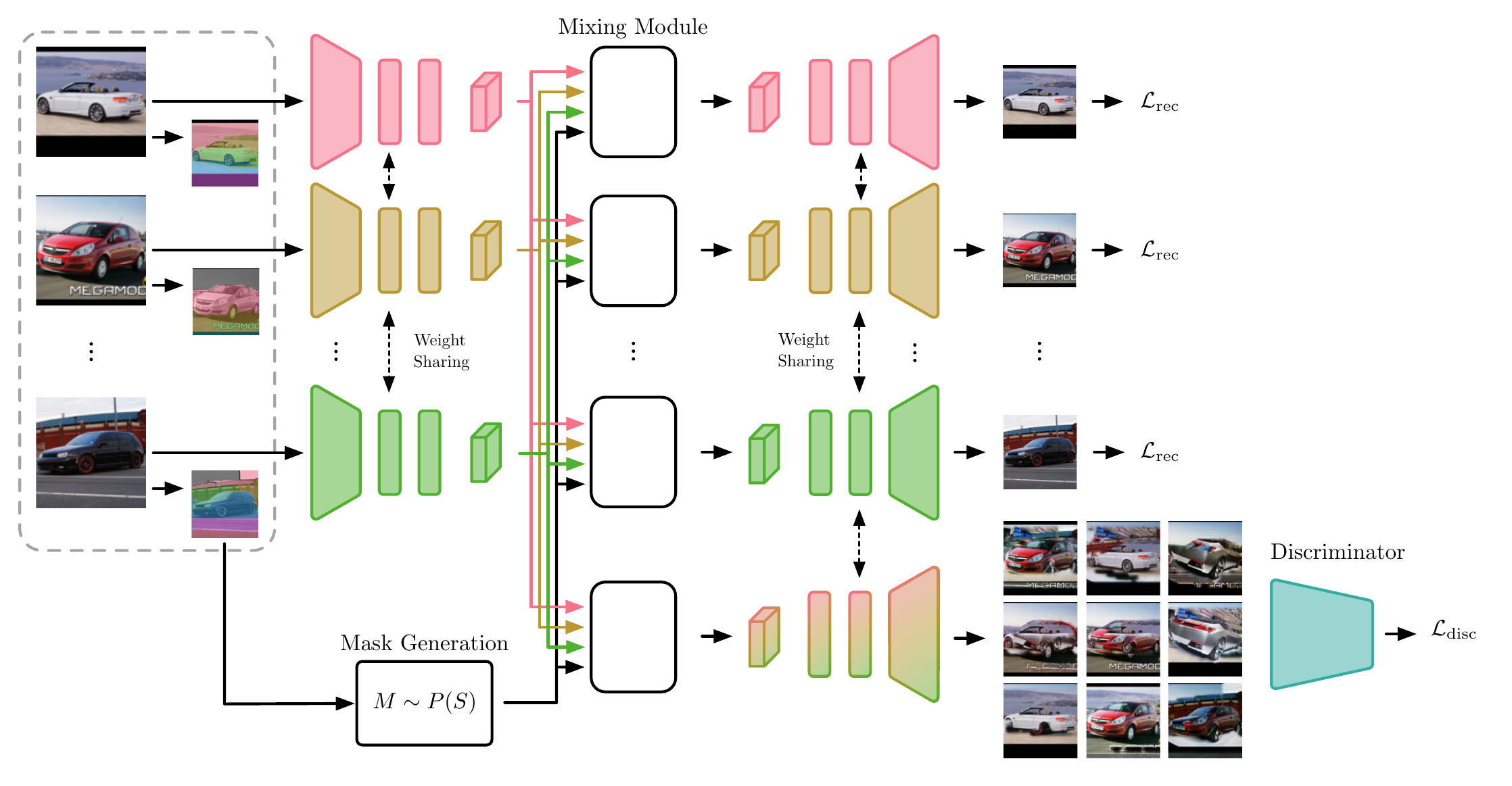}
    \caption{\methodname introduces a novel feature-mixing architecture that combines the content of an arbitrary number of images in a feature space guided by a segmentation-based mixing mask. The model is optimized with reconstruction and adversarial losses. Afterward, \methodname enables the generation of a large variety of new image compositions by sampling images and mixing masks.}
    \label{fig_overview}
\end{figure*}

\methodname advances the ChimeraMix pipeline to its full combinatorial potential and introduces a feature-mixing architecture for combining an arbitrary number of images guided by a segmentation-based mixing mask. Our new proposed method establishes a new state of the art in small data image classification.
Additionally, we present extensive ablation studies for crucial design choices of the pipeline and introduce a novel metric to assess the quality of data augmentation methods. 
An overview of \methodname is shown in \cref{fig_overview}.
In summary, our \textbf{contributions} in this work are:
\begin{itemize}
    \item A novel architecture for mixing the content of an arbitrary number of images to generate new image compositions.
    \item \methodname enables state-of-the-art image classification performance on small datasets.
    \item Enhancement of the mixing protocol for better sample quality.
    \item A new CLIP Synset Entropy evaluation to analyze the generalization capabilities of the generated images when using \methodname.
    \item Extensive analysis of the primary hyperparameters and ablation studies.
\end{itemize}

\section{Related Work}

Training deep neural networks with only a handful of samples requires extensive regularization to prevent overfitting.
Much research has already been put into the problem through general regularization methods, data augmentation, and, in particular, mixing pipelines.
This section gives a brief overview of the three areas of research and provides a context of related work for our method.

\subsection{Regularization for Small Data Problems}

Small data problems pose a unique challenge where each part of the standard machine learning pipeline has to be adapted.
\citet{barzDeepLearningSmall2020} propose to use a cosine loss function instead of the commonly used cross-entropy loss to prevent the logits for one class from dominating the others.
The convolutional neural tangent kernel (CNTK) has been used in \citep{aroraHarnessingPowerInfinitely2020} to study the behavior of ResNets on small datasets.
Other models, such as Random Forests \citep{breimanRandomForests2001}, are less prone to overfitting on small datasets.
This characteristic has been used by \cite{reindersObjectRecognitionVery2018} by combining them with neural networks to learn robust classifiers.
Finally, Wavelet filters \citep{vetterliWaveletsFilterBanks1992} have also been proposed in \citep{gauthierParametricScatteringNetworks2021} to classify images given only a handful of samples.

Since we are in the small data regime without any pretraining, we can not rely on models such as StableDiffusion \citep{rombachHighResolutionImageSynthesis2022} etc. that produce novel samples.
Approaches such as GLICO \citep{azuriGenerativeLatentImplicit2021} use a Generative Adversarial Network (GAN) to synthesize novel images, but this task of modeling the data distribution is aggravated by the restriction to only a few samples.
Instead, the data augmentation acts as a regularization for the model.

\subsection{Data Augmentation}

Synthetic samples can be created from a given set of images by randomly transforming each image.
Random cropping and vertical or horizontal flipping are commonly used to prevent the model from focusing on a specific absolute location of a feature in an image \citep{krizhevskyImageNetClassificationDeep2012}.
Other methods such as Cutout \citep{devriesImprovedRegularizationConvolutional2017} or RandomErasing \citep{zhongRandomErasingData2020} replace parts of the image with zeros or random noise.
The specific hyperparameters of the data augmentation pipeline can be finetuned to a dataset as with AutoAugment \citep{cubukAutoAugmentLearningAugmentation2019} or sampled randomly, such as with the TrivialAugment \citep{mullerTrivialAugmentTuningfreeStateoftheart2021} method.
Large models profit most from synthetically increasing the dataset size as they are more prone to overfitting \citep{brigatoCloseLookDeep2021,bornscheinSmallDataBig2020}.

One natural way of extending the size of a small image dataset is the combination or mixing of two or more images to create novel variations of the data.

\subsection{Mixing Augmentation}

A central method that uses multiple image-label pairs to create convex combinations of them is MixUp \citep{zhangMixupEmpiricalRisk2018}.
The method has been extended to mix the samples in a learned feature space in \citep{vermaManifoldMixupBetter2019}.
Since not all regions are equally suitable to be mixed, there are several methods that use the saliency of another model to guide the mixing process \citep{jeongSmoothMixTrainingConfidencecalibrated2021,kimPuzzleMixExploiting2020,daboueiSuperMixSupervisingMixing2021,kangGuidedMixupEfficientMixup2023}.
Another commonly used method is CutMix \citep{yunCutMixRegularizationStrategy2019}, where a rectangular region in one image is replaced by the content of another image from the same class.
\citet{liangMiAMixEnhancingImage2023} propose a system to apply mixing different transforms randomly that is similar to the random image transformations.
For a comprehensive overview and benchmark of different mixing augmentations, see \citet{liOpenMixupComprehensiveMixup2023}.
In contrast to the presented methods above, we propose a pipeline based on the simple principle of combining multiple samples with a powerful generator that is able to create novel samples.

\section{Method}

In this work, we present \methodname, a method for generating new image compositions from an arbitrary number of images in the small data regime, and establish a new state of the art on a set of datasets.
Our method builds on the ChimeraMix \citep{reindersChimeraMixImageClassification2022} approach and generalizes it to an arbitrary number of images.

The goal of our method is the generation of novel images by combining the semantic content of multiple images. 
Its main structure follows an encoder-decoder architecture similar to the CycleGAN architecture \citep{zhuUnpairedImagetoImageTranslation2017} that is trained adversarially using a discriminator.
An overview of \methodname is shown in Figure \ref{fig_overview}.
In the following sections, the model architectures of the generator and discriminator, the mixing module, and the training are presented.

\subsection{Encoder}

While techniques such as MixUp \citep{zhangMixupEmpiricalRisk2018} work in pixel space, the experiments in \citep{reindersChimeraMixImageClassification2022} have shown that it is beneficial to mix the images on a more abstract level in a learned feature space.
Thus, the first component of the \methodname generator is a convolutional encoder that maps the set of $N$ images $\ts{I} \in \mathbb{R}^{N \times C_\text{in} \times H \times W}$ with $C_\text{in}$ channels and a size of $H \times W$ to a lower dimensional feature map $\ts{F} \in \mathbb{R}^{N \times C_\text{f} \times H^{\prime} \times W^{\prime}}$ with $C_\text{f}$ channels and a size of $H^{\prime} \times W^{\prime}$. For the sake of simplicity, we omit the index for the batch.
The encoder consists of a block followed by $N_{\text{down}}$ convolutional blocks with a stride of $2$, which downsample the features and increase the number of features.
Each convolutional block consists of a 2D convolution, normalization, and ReLU activation.
In all our normalization layers, we use instance normalization to align the features adaptively.
Finally, the encoder has $N_{\text{blocks,enc}}$ residual blocks.
Note that this feature map retains some of the spatial information in order to allow for the subsequent mixing.

\subsection{Mixing Module}

The features are mixed guided by a mask so that each spatial region is filled with the features of a different image.
\methodname uses a segmentation-based mask generation where the images are segmented using the Felzenszwalb algorithm \citep{felzenszwalbEfficientGraphBasedImage2004}.
The resulting segmentations for all images are represented as $\ts{S} \in \mathbb{Z}^{N \times H \times W}$, where $S_{i,y,x}$ indicates the segmented region for each position.
The segmentations are downscaled to the size of the features
$\ts{S}^{\prime} \in \mathbb{Z}^{N \times H^{\prime} \times W^{\prime}}$ using nearest neighbor interpolation.

In the next step, we generate a mixing mask $\ts{M} \in [0, 1]^{N \times H^{\prime} \times W^{\prime}}$ with $\sum_{i = 1}^{N} \ts{M}_{i, h, w} = 1 \;\forall h \in [1, H^{\prime}], w \in [1, W^{\prime}]$. The mask is initialized with $\ts{M}_{n,h,w} = \delta_{n, 1}$, where $\delta$ is the Kronecker delta function. 
For each image $i \in [1, N]$, the set of all segments is defined as $\overline{S}_i = \{s \in \mathbb{Z} \mid \exists h \exists w : \ts{S}^{\prime}_{i,h,w} = s\}$. 
We randomly select segment regions from the current image by sampling a subset $\overline{S}_{i, \text{sampled}} \subseteq \overline{S}_i$, whereas each segment is included with a probability of $50\%$. The mask is updated to the selected segments by setting $M_{n,h,w} = \delta_{n, i}$ for all $h$ and $w$ if $\ts{S}^{\prime}_{i, h, w} \in \overline{S}_{i, \text{sampled}}$.

The mask values in our model are discrete, with each $M_{i,h,w}$ being an element of the set $\{0, 1\}$. We explored the possibility of generating continuous mask values by introducing a blend factor for each image. However, this approach was ultimately less effective than discrete masks, as the latter yielded superior performance in our evaluations.

There are several ways of generating the mixing masks.
In preliminary experiments, we determined that the segmentation-based sampling method shows better results than the grid-sampling method (\chimeramixgrid) when applied to more than two images.
Thus, after generating the \emph{segmentation}-based mixing mask, we calculate the mixed feature $\ts{F}_{\text{mix}} \in \mathbb{R}^{C_{\text{f}} \times H^{\prime} \times W^{\prime}}$ that combines the features of $N$ images as follows
\begin{equation}
    \ts{F}_{\text{mix}} = \sum_{i = 1}^{N} \ts{F}_{i} \odot \ts{M}_{i},
\end{equation}
where $\odot$ is the element-wise multiplication.

\subsection{Decoder}

Finally, the mixed features are passed to a convolutional decoder that maps from the feature space back to the pixel space. The design of the decoder is asymmetric to the encoder. The decoder consists of $N_{\text{blocks,dec}}$ residual blocks followed by $N_{\text{down}}$ transposed convolution blocks that upscale the features. At the end, a projection layer transforms the features to the image dimension $C_\text{in}$, producing the generated image $\hat{\ts{I}} \in \mathbb{R}^{C_\text{in} \times H \times W}$.

\subsection{Discriminator}

The discriminator receives the original and generated images as input and has the task of distinguishing the real from the fake images.
The discriminator network has four convolutional blocks for extracting features. Each convolutional block consists of a 2D convolutional layer, normalization layer, and LeakyReLU activation. The convolutional layers, except for the first layer, have a stride of two to downscale the features.
Afterward, an output convolutional layer projects the features to a one-channel feature map followed by a Sigmoid activation.
The generator and discriminator are trained adversarially, which will be explained in the next section.

\begin{figure}[t]
    \centering
    \includegraphics[width=0.99\linewidth]{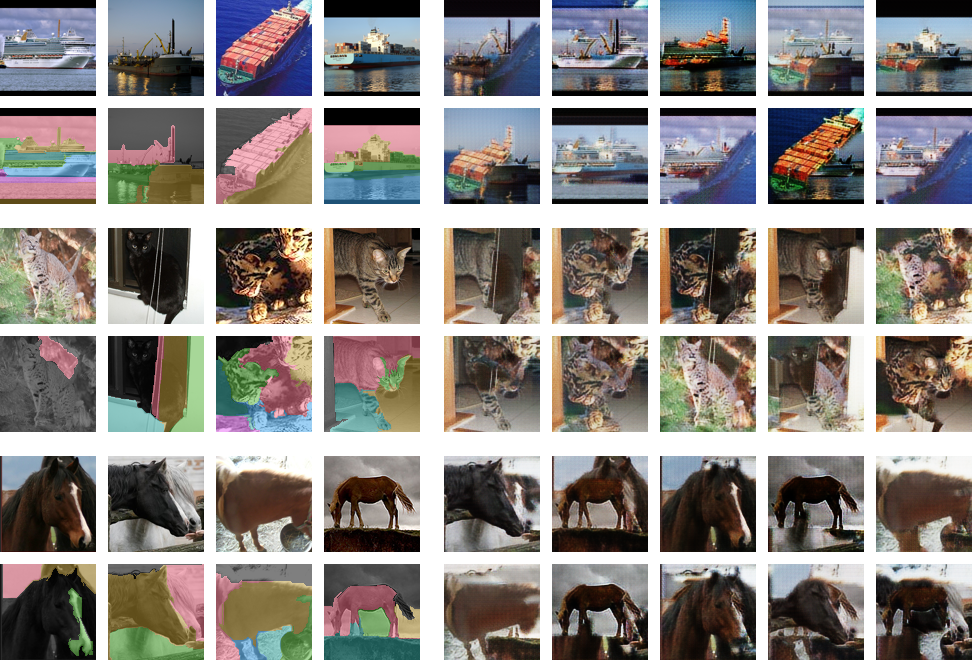}
    \caption{Qualitative samples using the \methodname method on three classes from the \stl dataset. For each class, four original images and their segmentation masks (left) and ten generated image compositions (right) are shown.}
    \label{fig_mix_visualization}
\end{figure}

\subsection{Training}
\label{sec_method_training}

The generator $\operatorname{Gen}(\ts{I}, \ts{M})$ combines the encoder $\operatorname{Enc}$, mixing module $\operatorname{Mix}$, and decoder $\operatorname{Dec}$ to produce the generated image $\hat{\ts{I}}$,
\begin{equation}
    \hat{\ts{I}} = \operatorname{Gen}(\ts{I}, \ts{M})= \operatorname{Dec}(\operatorname{Mix}(\operatorname{Enc}(\ts{I}), \ts{M})),
\end{equation}
based on the input images $\ts{I}$ and mixing mask $\ts{M}$.

To enable the generation of novel images from only a handful of samples, \methodname uses multiple loss terms.
When the mixing module receives a mask $\ts{M}^{\text{rec},i}$ which selects only the features of one of the images, i.e., $\ts{M}^{\text{rec},i}_{j,h,w} = \delta_{j,i}$, a reconstruction loss and a perceptual loss are applied. The reconstruction loss $\mathcal{L}_{\text{rec}}$ averages the mean squared error between the original image and the generated images over all $N$ images with the respective mask $\ts{M}^{\text{rec},i}$, i.e., 
\begin{equation}
    \mathcal{L}_{\text{rec}} = \frac{1}{NCHW}\sum_{i=1}^N \norm{\operatorname{Gen}(\ts{I}, \ts{M}^{\text{rec},i}) - \ts{I}_i}^2_2.
\end{equation}

\begin{table*}[ht!]
\centering
\small
\caption{Test accuracy on \cifairX, \stl, and \cifairC with different numbers of samples per class for training comparing a standard classification baseline \citep{zagoruykoWideResidualNetworks2016,heDeepResidualLearning2016} as well as state-of-the-art methods for small data image classification, i.e., Cutout \citep{devriesImprovedRegularizationConvolutional2017}, Random Erasing \citep{zhongRandomErasingData2020}, Cosine \citep{barzDeepLearningSmall2020}, MixUp \citep{zhangMixupEmpiricalRisk2018}, MixUpN\dag \citep{zhangMixupEmpiricalRisk2018}, Scattering \citep{gauthierParametricScatteringNetworks2021}, GLICO \citep{azuriGenerativeLatentImplicit2021}, ChimeraMix+Grid \citep{reindersChimeraMixImageClassification2022}, ChimeraMix+Seg \citep{reindersChimeraMixImageClassification2022}, and SuperMix \citep{daboueiSuperMixSupervisingMixing2021}. The best result is highlighted in bold. 
Note, that the size of the dataset of \cifairC with \num{5} samples per class is the same as that of \cifairX and \stl with \num{50} samples. 
\dag MixUpN is an extension of MixUp to multiple images by sampling the weighting from a Dirichlet distribution.
}
\label{tab_results_sota}
\begin{tabular}{ll
S[table-format=2.2, table-space-text-post=\std{9.99}]
S[table-format=2.2, table-space-text-post=\std{9.99}]
S[table-format=2.2, table-space-text-post=\std{9.99}]
S[table-format=2.2, table-space-text-post=\std{9.99}]
S[table-format=2.2, table-space-text-post=\std{9.99}]
S[table-format=2.2, table-space-text-post=\std{9.99}]}
\toprule
\multicolumn{2}{l}{Samples per Class} & {5} & {10} & {20} & {30} & {50} & {100} \\
{Dataset} & {Method} & {} & {} & {} & {} & {} & {} \\
\midrule
\multirow[c]{10}{*}{ciFAIR-10} & Baseline & 31.37\std{3.28}{} & 38.09\std{1.34}{} & 47.50\std{2.09}{} & 53.19\std{0.60}{} & 58.84\std{0.82}{} & 70.34\std{1.17}{} \\
 & Cutout & 28.88\std{2.84}{} & 37.33\std{1.03}{} & 47.55\std{2.06}{} & 53.39\std{1.32}{} & 61.17\std{1.03}{} & 72.14\std{1.10}{} \\
 & Random Erasing & 28.91\std{2.64}{} & 37.13\std{0.61}{} & 47.20\std{2.32}{} & 53.11\std{1.65}{} & 60.34\std{0.35}{} & 72.00\std{0.71}{} \\
 & Cosine & 31.45\std{3.22}{} & 37.88\std{1.24}{} & 46.69\std{1.38}{} & 52.16\std{0.72}{} & 59.24\std{1.60}{} & 70.18\std{1.32}{} \\
 & MixUp & 33.41\std{2.70}{} & 43.03\std{1.21}{} & 53.09\std{1.00}{} & 59.47\std{1.10}{} & 66.16\std{0.78}{} & 74.23\std{0.35}{} \\
 & MixUpN\dag & 32.83\std{2.51}{} & 41.92\std{1.19}{} & 53.16\std{0.59}{} & 58.53\std{0.36}{} & 64.88\std{0.68}{} & 73.19\std{0.75}{}\\
 & Scattering  & 30.50\std{3.87}{} & 37.28\std{1.87}{} & 45.65\std{1.45}{} & 50.47\std{1.19}{} & 54.30\std{0.95}{} & 61.51\std{0.79}{} \\
 & GLICO & 31.91\std{2.41}{} & 42.02\std{0.87}{} & 51.61\std{1.23}{} & 59.03\std{0.70}{} & 65.00\std{1.24}{} & 73.96\std{0.81}{} \\
 & ChimeraMix+Grid & 36.94\std{2.63}{} & 45.57\std{2.11}{} & 53.67\std{2.84}{} & 59.66\std{1.35}{} & 65.42\std{0.83}{} & 73.76\std{0.30}{} \\
 & ChimeraMix+Seg &  37.31\std{2.57}{} &  47.60\std{1.81}{} &  56.21\std{1.77}{} &  60.92\std{0.62}{} &  67.30\std{1.21}{} &  74.96\std{0.21}{} \\
 & \methodname & \textbf{39.05}\std{2.77}{} & \textbf{48.84}\std{2.28}{} & \textbf{57.56}\std{1.36}{} & \textbf{63.22}\std{0.53}{} & \textbf{68.95}\std{0.85}{} & \textbf{76.24}\std{0.49}{}  \\
\midrule
\multirow[c]{9}{*}{STL-10} & Baseline & 27.61\std{0.90}{} & 31.93\std{1.68}{} & 36.50\std{0.94}{} & 39.95\std{1.26}{} & 44.82\std{0.48}{} & 53.51\std{1.65}{} \\
 & Cutout  & 28.05\std{1.73}{} & 31.45\std{2.46}{} & 37.68\std{1.30}{} & 40.69\std{1.13}{} & 45.63\std{1.19}{} & 54.32\std{1.01}{} \\
 & Random Erasing & 27.87\std{1.36}{} & 31.32\std{0.48}{} & 36.91\std{1.45}{} & 40.66\std{0.84}{} & 45.93\std{1.10}{} & 53.31\std{1.52}{} \\
 & Cosine & 25.97\std{0.93}{} & 30.37\std{1.34}{} & 35.51\std{0.95}{} & 40.05\std{1.01}{} & 45.51\std{1.23}{} & 53.01\std{1.09}{} \\
 & MixUp & 30.06\std{1.80}{} & 35.63\std{0.85}{} & 42.44\std{1.85}{} & 45.00\std{2.71}{} & 49.03\std{1.34}{} & 54.38\std{2.11}{} \\
 & MixUpN\dag  & 28.93\std{2.54}{} & 34.45\std{1.04}{} & 41.31\std{0.63}{} & 42.17\std{2.09}{} & 46.21\std{2.27}{} & 49.57\std{1.15}{}\\
 & GLICO & 26.97\std{0.98}{} & 33.02\std{1.07}{} & 37.88\std{1.22}{} & 42.66\std{0.66}{} & 48.40\std{0.72}{} & 54.82\std{1.94}{} \\
 & ChimeraMix+Grid &  32.18\std{0.90}{} & 37.01\std{0.84}{} & 43.19\std{1.03}{} & 48.93\std{1.34}{} & 52.81\std{1.45}{} & 60.04\std{0.27}{} \\
 & ChimeraMix+Seg & 31.37\std{1.72}{} &  37.05\std{1.09}{} &  44.74\std{0.60}{} &  49.58\std{0.49}{} &  \textbf{55.06}\std{1.11}{} &  60.44\std{0.71}{} \\
 & \methodname & \textbf{33.09}\std{1.59}{} & \textbf{39.20}\std{0.76}{} & \textbf{45.80}\std{1.66}{} & \textbf{50.30}\std{0.99}{} & 54.10\std{1.22}{} & \textbf{60.87}\std{1.55}{} \\
\midrule
\multirow[c]{11}{*}{ciFAIR-100} & Baseline & 18.78\std{0.79}{} & 24.53\std{0.28}{} & 39.27\std{0.31}{} & 45.99\std{0.32}{} & 53.40\std{0.36}{} & 61.81\std{0.41}{} \\
 & Cutout & 19.25\std{0.52}{} & 27.77\std{0.39}{} & 40.72\std{0.68}{} & 47.78\std{0.39}{} & 55.13\std{0.30}{} &  63.26\std{0.62}{} \\
 & Random Erasing & 18.35\std{0.37}{} & 26.09\std{0.74}{} & 38.83\std{1.01}{} & 46.14\std{0.38}{} & 54.26\std{0.08}{} & 63.24\std{0.50}{} \\
 & Cosine & 18.04\std{0.87}{} & 23.72\std{0.35}{} & 38.84\std{0.73}{} & 45.83\std{0.43}{} & 53.32\std{0.11}{} & 61.50\std{0.46}{} \\
 & MixUp & 20.63\std{0.16}{} & 31.03\std{0.54}{} & 41.58\std{0.40}{} & 47.88\std{0.45}{} & 54.87\std{0.20}{} & 62.49\std{0.52}{} \\
 & MixUpN\dag & 17.71\std{0.51}{} & 26.36\std{0.49}{} & 37.54\std{0.65}{} & 44.71\std{0.67}{} & 52.54\std{0.44}{} & 58.16\std{0.42}{}\\
 & Scattering & 12.67\std{0.40}{} & 18.25\std{0.56}{} & 26.37\std{0.63}{} & 31.51\std{0.28}{} & 36.49\std{0.42}{} & 48.18\std{0.33}{} \\
 & GLICO & 19.32\std{0.39}{} & 28.49\std{0.60}{} & 40.45\std{0.30}{} & 45.90\std{0.77}{} & 53.53\std{0.19}{} & 60.68\std{0.50}{} \\
 & SuperMix & 19.23\std{0.45}{} & 26.78\std{0.20}{} & 38.47\std{0.83}{} & 44.69\std{0.63}{} & 53.07\std{0.13}{} & 62.63\std{0.30}{} \\
 & ChimeraMix+Grid & 20.24\std{0.12}{} & 31.62\std{0.82}{} & 41.80\std{0.52}{} & 48.10\std{0.71}{} & 54.67\std{1.01}{} & 62.13\std{0.27}{} \\
 & ChimeraMix+Seg &  21.09\std{0.47}{} &  32.72\std{0.60}{} &  43.23\std{0.38}{} &  48.83\std{0.72}{} &  55.79\std{0.21}{} & 62.96\std{0.77}{} \\
 & \methodname & \textbf{24.86}\std{0.54}{} & \textbf{33.80}\std{0.71}{} & \textbf{44.06}\std{0.64}{} & \textbf{49.46}\std{0.42}{} & \textbf{56.77}\std{0.33}{} & \textbf{63.91}\std{0.58}{} \\
\bottomrule
\end{tabular}%
\end{table*}

The visual appearance of the generated samples is improved by using a perceptual laplacian pyramid loss  $\mathcal{L}_{\text{per}}$ \citep{dentonDeepGenerativeImage2015}.
The laplacian pyramid $L^l(\ts{X})$ at level $l$ of an image $\ts{X}$ is defined as the difference between the image downsampled $l-1$ times and the image downsampled $l$ times and upsampled again once:
\begin{equation}
    L^l(\ts{X}) = \ts{X}\downarrow_{l-1} - \ts{X}\downarrow_{l}\uparrow,
\end{equation}
where $\downarrow$ is the downsampling operator and $\uparrow$ the upsample operator.
The downsample and upsample operators are both implemented using a convolution.
The laplacian pyramid loss is calculated by averaging the $l_1$ norm of the difference between the laplacian pyramid at level $l$ of the generated image and the original images over $L$ levels.
We compute the perceptual loss for two images per batch to reduce the computational requirements when training our method:
\begin{equation}
    \mathcal{L}_{\text{per}} = \frac{1}{N} \sum_{i=1}^N \sum_{l=1}^{L} \frac{1}{N_l}\norm{L^l(\operatorname{Gen}(\ts{I}, \ts{M}^{\text{rec},i})) - L^l(\ts{I}_i)}_1
\end{equation}
where $N_l$ is the number of elements of $L^l(\ts{I}_i)$.
This loss replaces the commonly used VGG \citep{simonyanVeryDeepConvolutional2015} feature losses that are common when more data or pretrained networks are available.

The generator and the discriminator are trained alternately. 
The discriminator is applied patch-wise and determines whether a patch is from a real or generated image by predicting a one or zero, respectively.
The generator, on the other hand, is trained to generate images that fool the discriminator by optimizing a loss $\mathcal{L}_{\text{G,disc}}$. The loss minimizes the distance between discriminator prediction for generated images and the real label.

Finally, the total generator loss is defined as follows:
\begin{equation}
    \label{eq:loss}
    \mathcal{L} = \alpha_{\text{rec}} \mathcal{L}_{\text{rec}} + \alpha_{\text{per}} \mathcal{L}_{\text{per}} + \alpha_{\text{G,disc}} \mathcal{L}_{\text{G,disc}},
\end{equation}
where $\alpha_{\text{rec}}$, $\alpha_{\text{per}}$, and $\alpha_{\text{G,disc}}$ are constants (see \cref{sec:hyperparameters}). 

Optionally, the generator operates on a higher image resolution by upscaling the input images $\ts{I}$ by bilinear interpolation. This does not add extra information but enables the generator to produce high-frequent image details more efficiently, which is especially useful for datasets with a small image size. The generated images are downscaled to the original image size before passing them to the classifier.
An analysis of the generator image size is shown in \cref{sec_analysis_generator_image_size}.

\begin{table*}[ht]
\caption{Accuracy of \chimeramixgrid, \chimeramixseg, and \methodname with AutoAugment (AA) and TrivialAugment (TA) on \cifairX, \stl, and \cifairC. The best result is highlighted in bold and the second best is underlined.}
\centering
\small
\label{tab_results_aa_ta}
\begin{tabular}{llcc
S[table-format=2.2, table-space-text-post=\std{9.99}]
S[table-format=2.2, table-space-text-post=\std{9.99}]
S[table-format=2.2, table-space-text-post=\std{9.99}]
S[table-format=2.2, table-space-text-post=\std{9.99}]
S[table-format=2.2, table-space-text-post=\std{9.99}]
S[table-format=2.2, table-space-text-post=\std{9.99}]
S[table-format=2.2, table-space-text-post=\std{9.99}]}
\toprule
\multicolumn{4}{l}{Samples per Class} & {5} & {10} & {30} & {50} & {100} \\
{Dataset} & {Method} & {AA} & {TA} & {} & {} & {} & {} & {} \\
\midrule
\multirow[c]{10}{*}{ciFAIR-10} & AutoAugment & \tick & \cross & 35.64\std{3.23}{} & 44.43\std{2.19}{} & 60.80\std{0.75}{} & 67.18\std{0.99}{} & 74.86\std{0.40}{} \\
 & TrivialAugment & \cross & \tick & 31.55\std{3.77}{} & 41.87\std{1.56}{} & 58.60\std{1.08}{} & 66.30\std{0.94}{} & 75.54\std{0.50}{} \\
 & ChimeraMix+Grid & \cross & \cross & 36.94\std{2.63}{} & 45.57\std{2.11}{} & 59.66\std{1.35}{} & 65.42\std{0.83}{} & 73.76\std{0.30}{} \\
 & ChimeraMix+Grid & \tick & \cross & 41.28\std{1.62}{} & 49.02\std{1.41}{} & 64.13\std{0.29}{} & 69.90\std{0.32}{} & 76.91\std{0.49}{} \\
 & ChimeraMix+Grid & \cross & \tick & 35.86\std{3.11}{} & 45.32\std{1.96}{} & 61.69\std{0.75}{} & 69.04\std{0.10}{} & 76.88\std{0.67}{} \\
 & ChimeraMix+Seg & \cross & \cross & 37.31\std{2.57}{} & 47.60\std{1.81}{} & 60.92\std{0.62}{} & 67.30\std{1.21}{} & 74.96\std{0.21}{} \\
 & ChimeraMix+Seg & \tick & \cross & \uline{42.16}\std{1.00}{} & 49.75\std{1.55}{} & 65.28\std{0.32}{} & 70.09\std{0.72}{} & 76.76\std{0.35}{} \\
 & ChimeraMix+Seg & \cross & \tick & 36.74\std{3.55}{} & 46.58\std{2.15}{} & 63.21\std{0.48}{} & 70.24\std{0.85}{} & \uline{77.79}\std{0.46}{} \\
 & \methodname & \cross & \cross & 39.05\std{2.77}{} & 48.84\std{2.28}{} & 63.22\std{0.53}{} & 68.95\std{0.85}{} & 76.24\std{0.49}{} \\
 & \methodname & \tick & \cross & \textbf{43.80}\std{2.53}{} & \textbf{52.66}\std{2.40}{} & \textbf{66.52}\std{0.27}{} & \uline{71.75}\std{0.70}{} & \textbf{77.96}\std{0.62}{} \\
 & \methodname & \cross & \tick & 42.06\std{3.23}{} & \uline{51.50}\std{2.40}{} & \uline{65.86}\std{0.46}{} & \textbf{71.86}\std{0.78}{} & 77.77\std{0.32}{} \\
\midrule
\multirow[c]{10}{*}{STL-10} & AutoAugment & \tick & \cross & 32.05\std{0.93}{} & 37.65\std{2.26}{} & 49.77\std{1.09}{} & 53.84\std{0.96}{} & 59.55\std{0.96}{} \\
 & TrivialAugment & \cross & \tick & 30.91\std{1.98}{} & 35.87\std{1.57}{} & 47.67\std{0.56}{} & 53.50\std{1.89}{} & 61.04\std{0.70}{} \\
 & ChimeraMix+Grid & \cross & \cross & 32.18\std{0.90}{} & 37.01\std{0.84}{} & 48.93\std{1.34}{} & 52.81\std{1.45}{} & 60.04\std{0.27}{} \\
 & ChimeraMix+Grid & \tick & \cross & \uline{37.54}\std{1.74}{} & 43.12\std{0.79}{} & 53.75\std{1.38}{} & 57.76\std{1.46}{} & 61.86\std{1.06}{} \\
 & ChimeraMix+Grid & \cross & \tick & 34.88\std{1.97}{} & 41.02\std{0.60}{} & 51.86\std{1.02}{} & 56.61\std{0.76}{} & 62.43\std{0.11}{} \\
 & ChimeraMix+Seg & \cross & \cross & 31.37\std{1.72}{} & 37.05\std{1.09}{} & 49.58\std{0.49}{} & 55.06\std{1.11}{} & 60.44\std{0.71}{} \\
 & ChimeraMix+Seg & \tick & \cross & 36.71\std{1.46}{} & \uline{43.88}\std{0.73}{} & \uline{54.90}\std{1.08}{} & 56.41\std{2.13}{} & 60.98\std{0.96}{} \\
 & ChimeraMix+Seg & \cross & \tick & 34.53\std{2.01}{} & 41.08\std{0.44}{} & 52.03\std{1.80}{} & 55.66\std{0.72}{} & \textbf{63.83}\std{0.52}{} \\
 & \methodname & \cross & \cross & 33.09\std{1.59}{} & 39.20\std{0.76}{} & 50.30\std{0.99}{} & 54.10\std{1.22}{} & 60.87\std{1.55}{}\\
 & \methodname & \tick & \cross & \textbf{37.87}\std{1.50}{} & 43.81\std{1.51}{} & \textbf{54.97}\std{0.74}{} & \textbf{58.39}\std{1.10}{} & \uline{63.82}\std{0.70}{} \\
 & \methodname & \cross & \tick & 37.21\std{1.81}{} & \textbf{44.16}\std{1.40}{} & 54.53\std{0.80}{} & \textbf{58.39}\std{0.84}{} & 62.87\std{2.50}{} \\
\midrule
\multirow[c]{10}{*}{ciFAIR-100}  & AutoAugment & \tick & \cross & 21.39\std{0.95}{} & 29.56\std{0.68}{} & 47.58\std{0.56}{} & 55.01\std{0.24}{} & 63.69\std{0.42}{} \\
 & TrivialAugment & \cross & \tick & 23.85\std{0.60}{} & 32.48\std{0.34}{} & 50.17\std{0.26}{} & 56.27\std{0.19}{} & 64.02\std{0.18}{} \\
 & ChimeraMix+Grid & \cross & \cross & 20.24\std{0.12}{} & 31.62\std{0.82}{} & 48.10\std{0.71}{} & 54.67\std{1.01}{} & 62.13\std{0.27}{} \\
 & ChimeraMix+Grid & \tick & \cross & 25.24\std{1.02}{} & 34.60\std{0.47}{} & 51.00\std{0.87}{} & 57.74\std{0.51}{} & 64.19\std{0.68}{} \\
 & ChimeraMix+Grid & \cross & \tick & 25.69\std{0.37}{} & 34.67\std{0.51}{} & 51.81\std{0.11}{} & 57.80\std{0.62}{} & 64.21\std{0.37}{} \\
 & ChimeraMix+Seg & \cross & \cross & 21.09\std{0.47}{} & 32.72\std{0.60}{} & 48.83\std{0.72}{} & 55.79\std{0.21}{} & 62.96\std{0.77}{} \\
 & ChimeraMix+Seg & \tick & \cross & 25.16\std{0.37}{} & 35.02\std{0.55}{} & 51.25\std{0.67}{} & 57.86\std{0.41}{} & \uline{64.39}\std{0.43}{} \\
 & ChimeraMix+Seg & \cross & \tick & \uline{26.36}\std{0.17}{} & \textbf{36.02}\std{0.22}{} & \textbf{52.74}\std{0.20}{} & \textbf{58.90}\std{0.64}{} & \textbf{64.79}\std{0.06}{} \\
 & \methodname & \cross & \cross & 24.86\std{0.54}{} & 33.80\std{0.71}{} & 49.46\std{0.42}{} & 56.77\std{0.33}{} & 63.91\std{0.58}{} \\
 & \methodname & \tick & \cross & \textbf{27.06}\std{0.72}{} & \uline{35.74}\std{0.88}{} & \uline{51.96}\std{0.28}{} & \uline{58.36}\std{0.48}{} & 63.85\std{0.13}{} \\
 & \methodname & \cross & \tick & 26.21\std{0.58}{} & 35.23\std{0.89}{} & 51.45\std{0.44}{} & 58.11\std{0.62}{} & 64.33\std{0.26}{} \\
\bottomrule
\end{tabular}
\end{table*}

\section{Experiments}

To evaluate the performance of the proposed method, we perform several experiments that test the impact of our generated images when training a classifier, the structure of the feature space, and an extensive set of ablation and sensitivity analyses of the crucial parts of the pipeline.

We evaluate our method on three benchmark datasets that are commonly used to evaluate classification algorithms in the small data regime.
The ciFAIR-10 and ciFAIR-100 datasets \citep{barzWeTrainTest2020} contain $\num{50000}$ and $\num{10000}$ $32\times32$ images in their training and test sets that are assigned to one of 10 and 100, respectively, classes.
We follow the standard procedure of subsampling these images further so that they only contain a handful of samples per class for training.
Additionally, we evaluate \methodname on the STL-10 dataset \citep{coatesAnalysisSinglelayerNetworks2011}, which offers more complex samples of size $96\times96$ in a $\num{5000}/\num{8000}$ training and test split.
This dataset is subsampled as well, similar to the other two.

\subsection{Experimental Setup}

\begin{figure}[t!]
  \centering
  \includegraphics[width=0.99\linewidth]{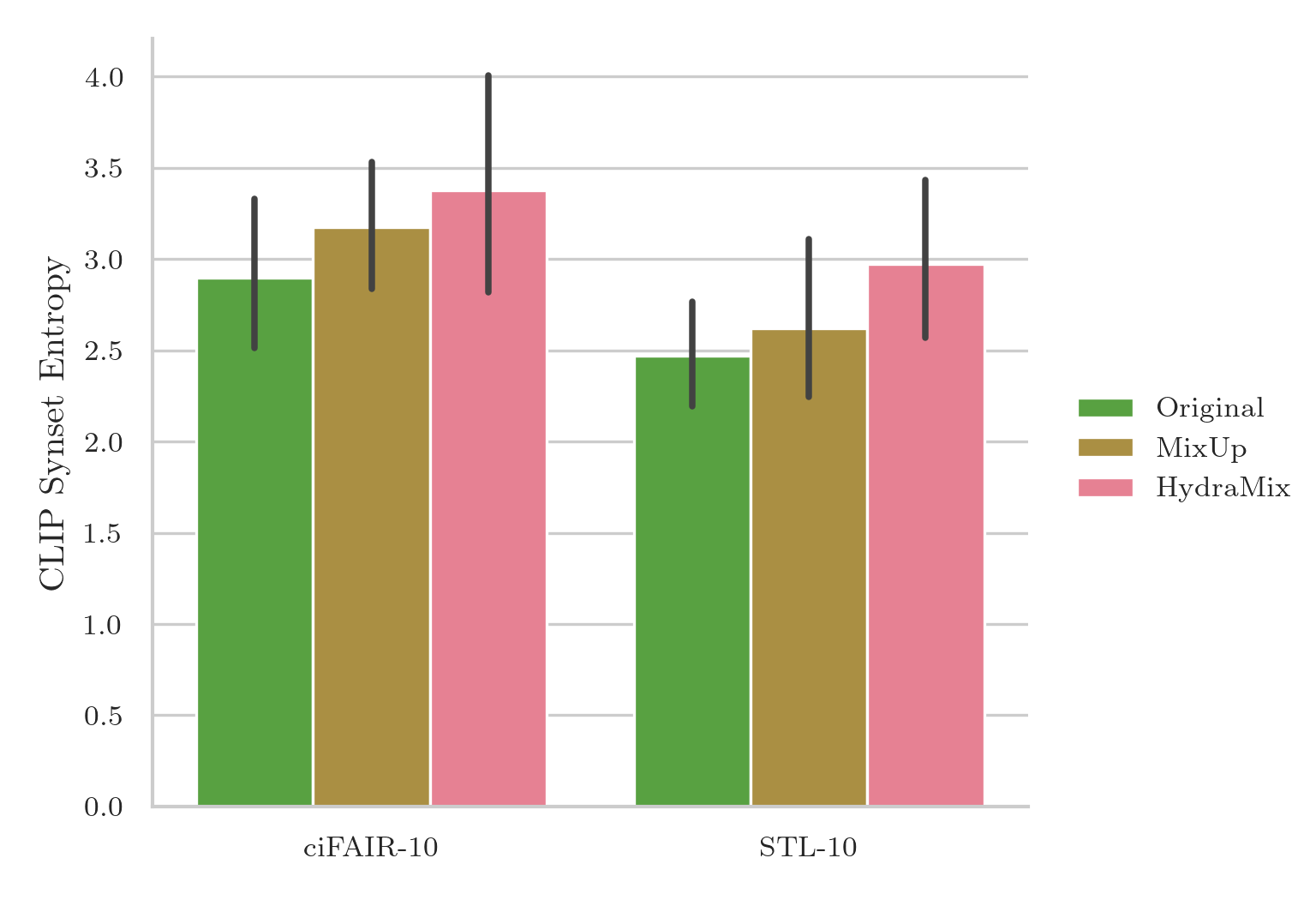}
  \caption{CLIP Synset Entropy ($\uparrow$) of the original dataset, a dataset generated with MixUp, and a dataset generated with \methodname. By sampling new compositions, \methodname is able to generate a larger variety of images that cover more synset concepts.} %
  \label{fig_clip_entropy_main}
\end{figure}

For the \cifairX and \cifairC datasets, we train a WideResNet-16-8 \citep{zagoruykoWideResidualNetworks2016} classifier using the same hyperparameters as \citep{brigatoTuneItDon2021}.
On \stl, we use a ResNet-50 \citep{heDeepResidualLearning2016} due to the larger images.
For all experiments, we use the same hyperparameters if not noted otherwise and optimize the networks using SGD with momentum and a cosine-annealing schedule for the learning rate.
We report average metrics over five different seeds together with their standard deviation.
The image segmentation is calculated in advance, such that the additional processing time is negligible.

We augment the dataset of the image classifier with generations from the trained \methodname pipeline by injecting the samples into the classification training pipeline.
Qualitative examples are shown in \cref{fig_mix_visualization}.
The results are reported for a pipeline that is trained to mix four images.
We determined this hyperparameter using a grid search and report the results in \cref{sec:mixing_multiple_images}.
In our experiments in \cref{sec:analysis_mixing_ratio}, we also provide a sensitivity analysis for the ratio $p_{\text{gen}}$ between real and generated images and its effect on the final validation accuracy.

\subsection{Comparison with State of the Art}

\begin{figure*}[h!]
  \centering
  \includegraphics[width=\linewidth]{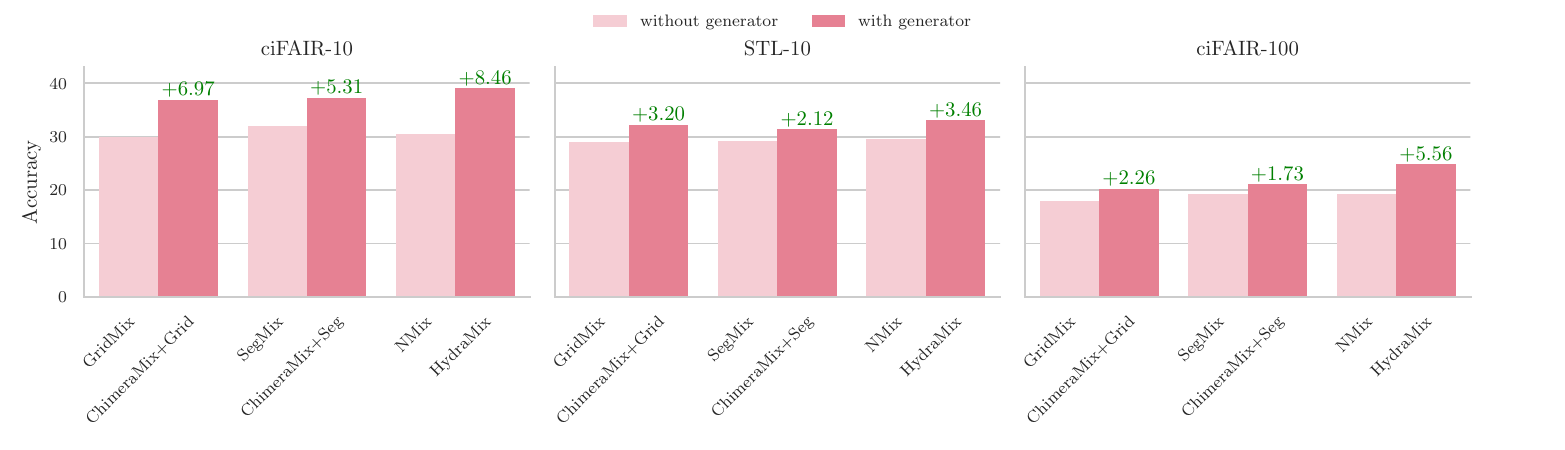}
  \caption{Average classification accuracy on the evaluated datasets with $5$ samples per class of \methodname, \chimeramixgrid, and \methodname and their respective ablation methods that do not use a generator.}
  \label{fig_ablation_generator_direct_nmix}
\end{figure*}

Following the methodology in \citep{reindersChimeraMixImageClassification2022}, we evaluate \methodname against an extensive set of state-of-the-art methods and baselines such as Cutout \citep{devriesImprovedRegularizationConvolutional2017}, Random Erasing \citep{zhongRandomErasingData2020}, Cosine loss \citep{barzDeepLearningSmall2020}, MixUp \citep{zhangMixupEmpiricalRisk2018}, MixUp extended to multiple images denoted as MixUpN, GLICO \citep{azuriGenerativeLatentImplicit2021}, \chimeramixgrid\citep{reindersChimeraMixImageClassification2022}, and \chimeramixseg\citep{reindersChimeraMixImageClassification2022}.
Additionally, on \cifairX and \cifairC, we compare our method with the Parametric Scattering Networks \citep{gauthierParametricScatteringNetworks2021} and on \cifairX we include the saliency-based method SuperMix \citep{daboueiSuperMixSupervisingMixing2021} in our evaluation. We also evaluate a WideResNet and ResNet-50, respectively, baseline for all settings.

The results of all methods are shown in \cref{tab_results_sota} on all datasets for different numbers of samples per class. On \cifairC with $5$ images per class, for example, the baseline achieves an accuracy of $31.37\%$, MixUp of $33.41\%$, \chimeramixseg of $37.31\%$, and \methodname of $39.05\%$.
On \stl with $10$ images per class, \methodname reaches $39.20\%$, outperforming \chimeramixseg and MixUp, which have a test accuracy of $37.05\%$ and $35.63\%$, respectively.
MixUpN is an extension of MixUp to multiple images by sampling the weighting from a Dirichlet distribution.
The method sometimes achieves results comparable to MixUp, but more frequently, it leads to decreased performance, often worse than the baseline, like on \cifairX and \cifairC with $5$ examples per class. This shows that a straightforward mixing of multiple images is not successful.

\begin{table}
    \caption{Comparison of different methods on ImageNet with $5$ examples per class.}
    \captionsetup{width=\linewidth} %

    \label{tab:results_imagenet}
    \centering
    \begin{tabular}{lS[table-format=1.2]S[table-format=2.2]}
    \toprule
    {Method} & {Top-1} & {Top-5} \\
    \midrule
    Baseline &  3.09 & 8.65 \\
    MixUp & 4.40 & 11.31 \\
    ChimeraMix & 9.44 & 20.76 \\
    HydraMix & \textbf{9.48} & \textbf{21.21} \\
    \bottomrule
    \end{tabular}
\end{table}

Furthermore, we evaluate the performance of a baseline, MixUp, \chimeramix, and \hydramix on \imagenet \citep{russakovskyImageNetLargeScale2015} with $5$ examples per class. The Top-1 and Top-5 accuracy are shown in \cref{tab:results_imagenet}. While the baseline reaches $3.09\%$, MixUp achieves $4.40\%$.
\chimeramix and \hydramix significantly outperform both methods and achieve an accuracy of $9.44\%$ and $9.48\%$, respectively.

Overall, the results demonstrate the superior performance of \methodname in comparison to all other methods.
Particularly in scenarios with limited data, such as $5$ or $10$ samples per class, the proposed method demonstrates a noticeable advantage over the prior state-of-the-art established by \chimeramixseg.

\begin{table}
    \caption{
    Self-supervised learning analysis: Accuracy of DINO and DINO with \chimeramix and \hydramix on STL-10 with $5$ training examples per class.}
    \label{tab:ablation_dino}
    \captionsetup{width=\linewidth} %

    \centering
    \begin{tabular}{lS[table-format=2.2, table-space-text-post=\std{9.99}]}
    \toprule
    {Method}  \\
    \midrule
    DINO &  30.81\std{4.30} \\
    DINO+ChimeraMix &  32.87\std{3.72} \\
    DINO+HydraMix &  \textbf{33.20}\std{3.74} \\
    \bottomrule
    \end{tabular}
\end{table}

\subsection{Automatic Augmentation}

In the next experiment, we evaluate the combination of our method with automatic augmentation techniques. While previous experiments use only basic augmentation techniques to assess the impact of the methods on the performance of the classifier, current image classification pipelines commonly use automatic augmentation policies to prevent overfitting.
The results of AutoAugment \citep{cubukAutoAugmentLearningAugmentation2019}, TrivialAugment \citep{mullerTrivialAugmentTuningfreeStateoftheart2021}, and combinations of \chimeramixgrid, \chimeramixseg, and \methodname with both automatic augmentation methods are shown in \cref{tab_results_aa_ta}.
The experiment demonstrates that \methodname is able to usually outperform AutoAugment and TrivialAugment, especially when few training examples are available. 
On \cifairX with $5$ examples per class, for example, AutoAugment achieves an accuracy of $35.64\%$ while \methodname reaches $39.05\%$. The combination of \methodname with AutoAugment boosts the performance to $43.89\%$.
However, AutoAugment is not strictly in the small data regime as its augmentation policy was finetuned on the whole dataset.

\subsection{Comparison to Self-supervised and Generative Methods}

Additionally, we evaluate a self-supervised learning approach and a state-of-the-art generative method. The experiments are performed on \stl with $5$ examples per class.
DINO \citep{caronEmergingPropertiesSelfSupervised2021} is a widely used self-supervised learning method that leverages contrastive learning to extract meaningful representations from unlabelled data, achieving state-of-the-art performance in various tasks.
We train DINO according to the original implementation and training protocol. Afterward, a linear probing is performed. The results are reported in \cref{tab:ablation_dino}.
DINO achieves an accuracy of $30.81\%$. When integrating \chimeramix and \hydramix, the performance is increased to $32.87\%$ and $33.20\%$, respectively.

Next, a comparison to a state-of-the-art generative method, DiffuseMix \citep{islamDiffuseMixLabelpreservingData2024}, is performed. DiffuseMix generates new images with a diffusion model, more precisely a InstructPix2Pix diffusion model \citep{brooksInstructpix2pixLearningFollow2023} based on Stable Diffusion with conditional prompts.
Afterward, a hybrid image is created and blended with a random fractal image.
It should be noted that the model was trained on the large-scale LAION-5B dataset, comprising over 5 billion images, and is therefore not in the small data regime.
The results in \cref{tab:ablation_diffusemix} show that DiffuseMix reaches $30.44\%$, while \hydramix outperforms it with an accuracy of $33.09\%$

\begin{table}
    \caption{Comparison of the generative augmentation method DiffuseMix with \chimeramix and \hydramix on STL-10 with $5$ training examples per class.}
    \label{tab:ablation_diffusemix}
    \captionsetup{width=\linewidth} %

    \centering
    \begin{tabular}{lS[table-format=2.2, table-space-text-post=\std{9.99}]}
    \toprule
    {Method}  \\
    \midrule
    Baseline & 27.61\std{0.90} \\
    DiffuseMix & 30.44\std{1.19} \\
    ChimeraMix+Grid &  32.18\std{0.90}{}   \\
    ChimeraMix+Seg & 31.37\std{1.72}{} \\
    HydraMix & \textbf{33.09}\std{1.59}{}  \\
    \bottomrule
    \end{tabular}
\end{table}

\subsection{CLIP Synset Entropy}
\label{sec_clip_synset_entropy}

\begin{table*}[ht]
\small
\centering
\caption{Analysis of the generator's impact. GridMix, SegMix, and MixN directly mix the images in pixel space without the generator of \chimeramix or \methodname. The study shows that mixing the features via the proposed generator (\chimeramixgrid, \chimeramixseg, and \methodname) is able to learn the generation of new image compositions and achieves a significantly improved performance.
}
\label{tab_ablation_generator_direct}
\begin{tabular}{llcc
S[table-format=2.2, table-space-text-post=\std{9.99}]
S[table-format=2.2, table-space-text-post=\std{9.99}]
S[table-format=2.2, table-space-text-post=\std{9.99}]
S[table-format=2.2, table-space-text-post=\std{9.99}]
S[table-format=2.2, table-space-text-post=\std{9.99}]
S[table-format=2.2, table-space-text-post=\std{9.99}]}
\toprule
\multicolumn{2}{l}{Samples per Class} & {5} & {10} & {20} & {30} & {50} & {100} \\
{Dataset} & {Method} & {} & {} & {} & {} & {} & {} \\
\midrule
\multirow[c]{6}{*}{ciFAIR-10} & GridMix & 29.97\std{1.17}{} & 39.90\std{1.24}{} & 48.60\std{3.18}{} & 54.99\std{2.49}{} & 61.12\std{1.59}{} & 72.41\std{0.69}{} \\
 & SegMix & 32.00\std{1.22}{} & 42.18\std{1.36}{} & 52.56\std{2.43}{} & 57.90\std{0.49}{} & 64.61\std{0.94}{} & 73.96\std{0.36}{} \\
 & NMix & 30.59\std{4.04}{} & 40.69\std{1.25}{} & 50.63\std{2.04}{} & 56.25\std{1.51}{} & 64.03\std{0.93}{} & 73.51\std{0.92}{}\\
 & ChimeraMix+Grid & 36.94\std{2.63}{} & 45.57\std{2.11}{} & 53.67\std{2.84}{} & 59.66\std{1.35}{} & 65.42\std{0.83}{} & 73.76\std{0.30}{} \\
 & ChimeraMix+Seg & 37.31\std{2.57}{} & 47.60\std{1.81}{} & 56.21\std{1.77}{} & 60.92\std{0.62}{} & 67.30\std{1.21}{} & 74.96\std{0.21}{} \\
 & \methodname & \textbf{39.05}\std{2.77}{} & \textbf{48.84}\std{2.28}{} & \textbf{57.56}\std{1.36}{} & \textbf{63.22}\std{0.53}{} & \textbf{68.95}\std{0.85}{} & \textbf{76.24}\std{0.49}{}\\
\midrule
\multirow[c]{6}{*}{STL-10} & GridMix & 28.98\std{1.49}{} & 31.21\std{1.52}{} & 37.08\std{1.09}{} & 42.14\std{1.52}{} & 49.33\std{0.88}{} & 56.92\std{0.51}{} \\
 & SegMix & 29.25\std{0.40}{} & 32.84\std{0.63}{} & 37.80\std{1.91}{} & 43.69\std{0.84}{} & 50.14\std{0.84}{} & 58.60\std{0.57}{} \\
 & NMix & 29.63\std{1.04}{} & 33.85\std{2.15}{} & 40.35\std{0.97}{} & 44.72\std{1.11}{} & 50.11\std{1.66}{} & 59.32\std{0.81}{}\\
& ChimeraMix+Grid & 32.18\std{0.90}{} & 37.01\std{0.84}{} & 43.19\std{1.03}{} & 48.93\std{1.34}{} & 52.81\std{1.45}{} & 60.04\std{0.27}{} \\
 & ChimeraMix+Seg & 31.37\std{1.72}{} & 37.05\std{1.09}{} & 44.74\std{0.60}{} & 49.58\std{0.49}{} & \textbf{55.06}\std{1.11}{} & 60.44\std{0.71}{} \\
 & \methodname & \textbf{33.09}\std{1.59}{} & \textbf{39.20}\std{0.76}{} & \textbf{45.80}\std{1.66}{} & \textbf{50.30}\std{0.99}{} & 54.10\std{1.22}{} & \textbf{60.87}\std{1.55}{} \\
\midrule
\multirow[c]{6}{*}{ciFAIR-100} & GridMix & 17.98\std{0.23}{} & 27.78\std{0.46}{} & 38.92\std{0.05}{} & 45.16\std{1.05}{} & 52.97\std{0.30}{} & 61.37\std{0.26}{} \\
 & SegMix & 19.36\std{0.82}{} & 29.62\std{0.22}{} & 41.00\std{0.42}{} & 47.50\std{0.38}{} & 54.62\std{0.15}{} & 62.43\std{0.38}{} \\
 & NMix & 19.30\std{0.31}{} & 29.30\std{0.71}{} & 39.90\std{0.53}{} & 46.26\std{0.30}{} & 54.13\std{0.32}{} & 62.85\std{0.54}{} \\
& ChimeraMix+Grid & 20.24\std{0.12}{} & 31.62\std{0.82}{} & 41.80\std{0.52}{} & 48.10\std{0.71}{} & 54.67\std{1.01}{} & 62.13\std{0.27}{} \\
 & ChimeraMix+Seg & 21.09\std{0.47}{} & 32.72\std{0.60}{} & 43.23\std{0.38}{} & 48.83\std{0.72}{} & 55.79\std{0.21}{} & 62.96\std{0.77}{} \\
 & \methodname & \textbf{24.86}\std{0.54}{} & \textbf{33.80}\std{0.71}{} & \textbf{44.06}\std{0.64}{} & \textbf{49.46}\std{0.42}{} & \textbf{56.77}\std{0.33}{} & \textbf{63.91}\std{0.58}{} \\
\bottomrule
\end{tabular}
\end{table*}

Analyzing the diversity of the generated data is very important. 
To show the generality of our approach, we employ the recently developed CLIP \citep{radfordLearningTransferableVisual2021} method in combination with the Open English WordNet \citep{millerWordNetLexicalDatabase1995,mccraeEnglishWordNet20202020} to measure the diversity of the generated data.
CLIP consists of a shared embedding space for images and natural language, which allows the computation of the similarity between an image and a given phrase.
Thus, it can act as an open-set classifier.
WordNet is a lexical database consisting of words that are connected by semantic relations.
Words are linked if they share the same meaning (these are called synonyms), if one is a more specific type of another (these are called hyponyms), or if they are parts of each other (known as meronyms).
For example, a hyponym of \texttt{vehicle} is \texttt{car} since a car is a type of vehicle.

The classification results indicate that \methodname generates images from a given set of examples that are more general than the original data or other augmentation methods.
To further support this, we map all dataset classes to synsets $s_i$ in the WordNet database, where $i \in [1, C]$ is the class index. For each synset, we extract all subordinate synsets that are hyponyms of $s_i$ denoted as $H_i = \{h_{i,1}, \dots, h_{i,N_i}\}$ and $N_i$ is the number of hyponyms. For example, the class \texttt{Dog} has the hyponyms \texttt{puppy}, \texttt{police dog}, and \texttt{Chihuahua} and \texttt{Truck} has the hyponyms \texttt{fire truck}, \texttt{tow car}, and \texttt{dump truck}, among others.
The features of the generated images should cover more subclasses of a given class.

We use a CLIP image encoder $f_{\text{Image}}^{\text{CLIP}}(\ts{x})$ and text encoder $f_{\text{Text}}^{\text{CLIP}}(t)$ based on a ViT-B/16 architecture \citep{dosovitskiyImageWorth16x162021} pretrained on LAION-2B \citep{schuhmannLAION5BOpenLargescale2022} using OpenCLIP \citep{ilharcoOpenCLIP2021}.
To encode synsets, we introduce a Synset-to-Text function $f_{\text{S2T}}(s)$ that generates a textual representation of a synset by combining all $K$ synonyms and the description of a synset in the format "$\textit{Synonym}_1, \dots, \textit{Synonym}_K\!: \textit{Description}$".
The similarity $\operatorname{sim}(\ts{x}, s)$ between an image $x$ and a synset $s$ is defined as the cosine similarity of the encoded image and the encoded synset:
\begin{equation}
    \operatorname{sim}(\ts{x}, s) = \frac{ f_{\text{Image}}^{\text{CLIP}}(\ts{x}) \cdot f_{\text{Text}}^{\text{CLIP}}(f_{\text{S2T}}(s))}{||f_{\text{Image}}^{\text{CLIP}}(\ts{x})|| \cdot||f_{\text{Text}}^{\text{CLIP}}(f_{\text{S2T}}(s))||}.
\end{equation}

\begin{table*}
\caption{Analysis of the number of mixing images for \methodname.}
\small
\centering
\label{tab_results_nmix}
\begin{tabular}{lc
S[table-format=2.2, table-space-text-post=\std{9.99}]
S[table-format=2.2, table-space-text-post=\std{9.99}]
S[table-format=2.2, table-space-text-post=\std{9.99}]
S[table-format=2.2, table-space-text-post=\std{9.99}]
S[table-format=2.2, table-space-text-post=\std{9.99}]
S[table-format=2.2, table-space-text-post=\std{9.99}]}
\toprule
\multicolumn{2}{l}{Samples per Class} & {5} & {10} & {20} & {30} & {50} & {100}  \\
{Dataset} & {\#Mixing Images} & {} & {} & {} \\
\midrule
\multirow[c]{4}{*}{ciFAIR-10} & 2 & 38.02\std{3.08}{} & 48.76\std{1.79}{} & 56.90\std{1.69}{} & 62.94\std{0.77}{} & 68.67\std{0.84}{} & 75.81\std{0.48}{} \\
 & 3 & 38.48\std{3.08}{} & 48.47\std{2.11}{} & \textbf{57.98}\std{1.45}{} & 62.97\std{0.89}{} & \textbf{69.12}\std{1.06}{} & 76.00\std{0.48}{} \\
 & 4 & \textbf{39.05}\std{2.77}{} & \textbf{48.84}\std{2.28}{} & 57.56\std{1.36}{} & 63.22\std{0.53}{} & 68.95\std{0.85}{} & \textbf{76.24}\std{0.49}{} \\
 & 5 & 37.84\std{3.43}{} & 48.19\std{1.73}{} & 57.75\std{1.98}{} & \textbf{63.39}\std{0.78}{} & 68.85\std{0.65}{} & 75.92\std{0.41}{} \\
\midrule
\multirow[c]{4}{*}{STL-10} & 2 & 32.32\std{1.69}{} & 38.57\std{1.62}{} & 44.76\std{1.44}{} & 49.28\std{1.50}{} & 54.31\std{0.85}{} & 61.24\std{1.12}{} \\
 & 3 & 32.90\std{2.15}{} & 38.36\std{1.38}{} & 44.83\std{0.24}{} & 49.75\std{1.07}{} & 54.62\std{0.47}{} & 60.99\std{1.25}{} \\
 & 4 & \textbf{33.09}\std{1.59}{} & 39.20\std{0.76}{} & 45.80\std{1.66}{} & 50.30\std{0.99}{} & 54.10\std{1.22}{} & 60.87\std{1.55}{} \\
 & 5 & 32.57\std{1.92}{} & \textbf{39.28}\std{1.57}{} & \textbf{46.10}\std{0.66}{} & \textbf{50.66}\std{0.71}{} & \textbf{55.48}\std{0.78}{} & \textbf{62.36}\std{0.94}{} \\
\midrule
\multirow[c]{4}{*}{ciFAIR-100} & 2 & 23.50\std{0.45}{} & 33.84\std{0.64}{} & 44.02\std{0.58}{} & 50.03\std{0.66}{} & \textbf{56.95}\std{0.33}{} & 64.24\std{0.26}{} \\
 & 3 & 24.66\std{0.77}{} & \textbf{34.03}\std{0.77}{} & \textbf{44.51}\std{0.77}{}  & \textbf{50.15}\std{0.76}{} & 56.62\std{0.44}{} & \textbf{64.41}\std{0.23}{} \\
 & 4 & \textbf{24.86}\std{0.54}{} & 33.80\std{0.71}{} & 44.06\std{0.64}{} & 49.46\std{0.42}{} & 56.77\std{0.33}{} & 63.91\std{0.58}{} \\
 & 5 & 24.58\std{0.85}{} & 33.82\std{0.41}{} & 43.88\std{0.69}{} & 49.62\std{0.45}{} & 56.48\std{0.47}{} & 63.88\std{0.31}{}  \\
\bottomrule
\end{tabular}
\end{table*}

We compute the CLIP embeddings of the original images, the images produced by the MixUp method, and our generated images. For MixUp and \methodname, we sample $1000$ images per class. Then, the synsets $h_{i,j}$ are encoded, and similarities between all images and synsets are calculated.
For each dataset, we calculate the synset distribution per class $P(h_{i,j})$ by averaging the synset distribution for each image, which is determined as the softmax across all synsets:
\begin{equation}
    P(h_{i,j}) = \frac{1}{N} \sum_{l=1}^{N}  \frac{\operatorname{exp}(\operatorname{sim}(\ts{x}_l, h_{i,j}) / \tau)}{\sum_{k=1}^{N_i} \operatorname{exp}(\operatorname{sim}(\ts{x}_l, h_{i,k}) / \tau)},
\end{equation}
where $\tau$ is a temperature parameter and $N$ the number of images per class.

Finally, we measure the diversity of the dataset per class by computing the entropy of the synset distribution:
\begin{equation}
    \operatorname{CSE}_i = -\sum_{j=1}^{N_i} P(h_{i,j}) \log(P(h_{i,j})).
\end{equation}
The CLIP Synset Entropy $\operatorname{CSE}_i$ measures the hyponym coverage of a set of images for each class.
We compute the overall CLIP Synset Entropy across all classes as $\operatorname{CSE}$ by averaging $\operatorname{CSE}_i$.

The results, shown in \cref{fig_clip_entropy_main}, support the outcome of our classification experiments.
The features of the images that are generated by \methodname are more similar to a larger range of synset concepts.
On both datasets, \methodname enables the generation of a more diverse range of images, which enhances the average synset similarity distribution.
On \stl, the original dataset has a $\operatorname{CSE}$ of $2.47$. MixUp achieves $2.61$ and \methodname reaches $2.97$.

\subsection{Generator Impact}

\begin{figure*}[ht]
  \centering
  \includegraphics[page=1, width=\linewidth]{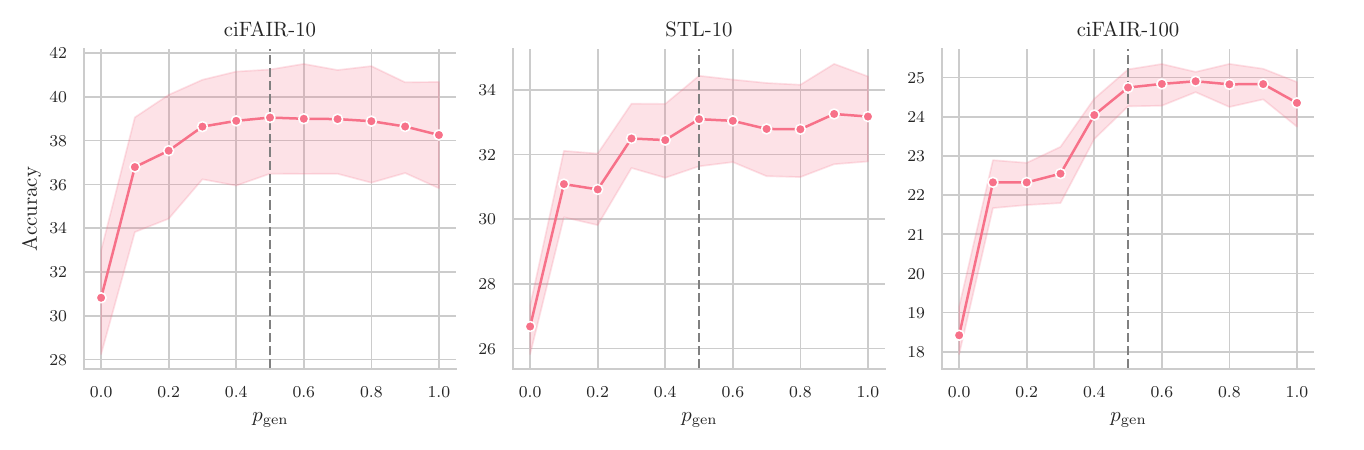}
  \caption{Evaluation of the mixing probability $p_\text{gen}$ on different datasets. The default mixing ratio of $0.5$ is indicated by a dashed line. Experiments are performed with 5 examples per class.}
  \label{fig:ablation_sample_ratio}
\end{figure*}

We investigate the impact of the proposed generator on the generated samples.
For this, we evaluate each method (\chimeramixgrid, \chimeramixseg, and \methodname) against a baseline that performs the mixing in pixel space without the use of the generator. The pixel-based methods will be denoted as GridMix, SegMix, and NMix. The average accuracy of all methods is shown in \cref{fig_ablation_generator_direct_nmix} with $5$ examples per class.
The results demonstrate a noticeable improvement for all methods that employ a generator on all datasets, highlighting the importance of the generator.
The most significant improvement is apparent for \methodname. On \cifairC, for example, the performance without the generator (NMix) is $30.59\%$, while the performance with the generator is $39.05\%$ ($+8.46$ percentage points).

The results in \cref{fig_ablation_generator_direct_nmix} demonstrate the advantage of the proposed generator and its operation in the mixed feature space.
We report detailed performance metrics for varying numbers of samples per class in \cref{tab_ablation_generator_direct}.
It is apparent that our pipeline is especially useful in the small data regime with 20 samples or less per class, but also increases performance with more samples.

\subsection{Object Detection}

\begin{table*}
    \caption{Object detection performance on COCO with very few training samples using Faster R-CNN when integrating \hydramix as a proof-of-concept.}
    \label{tab:results_object_detection}
    \centering
    \begin{tabular}{cl
        S[table-format=1.2, table-space-text-post={\std{0.12} (+23.3\%)}]
        S[table-format=1.2, table-space-text-post={\std{0.12} (+23.3\%)}]
        S[table-format=1.2, table-space-text-post={\std{0.12} (+23.3\%)}]}
    \toprule
    Shot & {Method} & {$\operatorname{AP}$} & {$\operatorname{AP_{50}}$} & $\operatorname{AP_{75}}$ \\
    \midrule
    \multirow{2}{*}{1} & Baseline & 1.27 \std{0.06} & 2.90 \std{0.20} & 1.00 \std{0.00} \\
    & Baseline + HydraMix & 1.63 \std{0.15} (+28.9\%) & 3.73 \std{0.32} (+28.7\%) & 1.23 \std{0.12} (+23.3\%) \\
    \midrule %
    \multirow{2}{*}{3} & Baseline & 3.37 \std{0.06} & 7.20 \std{0.20} & 2.83 \std{0.15} \\
    & Baseline + HydraMix & 4.03 \std{0.21} (+19.8\%) & 8.53 \std{0.31} (+18.5\%) & 3.40 \std{0.30} (+20.0\%) \\ 
    \bottomrule
    \end{tabular}
\end{table*}

Furthermore, we extend \hydramix to object detection tasks as a proof-of-concept, demonstrating its applicability beyond classification.
Using Faster R-CNN \citep{renFasterRCNNRealtime2015}, we evaluate the method on the COCO dataset \citep{linMicrosoftCOCOCommon2014}.
For training, we subsample the dataset by randomly selecting $N$ images for each class, denoted as $N$-shot.
\hydramix leverages the bounding box information to generate mixing masks.
The results, presented in \cref{tab:results_object_detection}, highlight the potential of \hydramix in enhancing performance in complex vision tasks.
On $1$-shot and $3$-shot, \hydramix boosts the performance by $28.9\%$ and $19.8\%$, respectively.
This approach also opens up significant opportunities for further extensions, particularly by incorporating semantic information.

\subsection{Analyses}

In the following, we perform several analyses to examine the robustness of the individual components as well as the parameters of the proposed method.

\subsubsection{Mixing Multiple Images}
\label{sec:mixing_multiple_images}

The presented method enables mixing more than two images, which exponentially expands the space of combinations. In the next experiment, we analyze the influence of mixing multiple images.
There is a trade-off between the number of segments in each image, which is heavily influenced by the image size, and the noise that mixing more than two images introduces into the data samples.
Thus, the optimal number of images varies and depends on the number of samples per class and the image size.
We perform a grid search over the number of mixed images between two and five for all datasets and evaluate the downstream classification performance.
The results in \cref{tab_results_nmix} demonstrate that mixing multiple images is very effective. The optimal number of images indeed varies depending on the dataset. 
In the case of \stl, the optimal performance is attained by mixing a larger quantity of images, specifically five images. For \cifairC, the best results are achieved by combining three images.
Overall, mixing four images consistently yields strong performance across all configurations of this hyperparameter.

\subsubsection{Mixing Ratio}
\label{sec:analysis_mixing_ratio}

The amount of the generated images is controlled by the mixing probability $p_{\text{gen}}$. At each training step of the classifier, we sample with $p_{\text{gen}}$ whether generating new images or training on a batch of original images.
A larger $p_{\text{gen}}$ results in a more extensive augmented dataset because the classifier is trained on more generated samples.
It controls the trade-off between data variability and the similarity of the training data distribution to the original small dataset.
In \cref{fig:ablation_sample_ratio}, we show the results of training a large number of classifiers with different mixing ratios in the setting with 5 samples per class and report the average validation accuracy. When the mixing ratio is set to zero, we only train with the original images.
When $p_{\text{gen}} = 1$, the classifier is trained with generated images exclusively.
This experiment shows that integrating a large portion of generated images (i.e., a mixing ratio between $40\%$ and $90\%$) is very effective. The optimal mixing ratio varies slightly between the datasets but is generally reasonably robust. 
Replacing every second image in the classifier pipeline with a generated mix is a good choice across all datasets.

\begin{figure}[t]
  \centering
  \includegraphics[width=\linewidth]{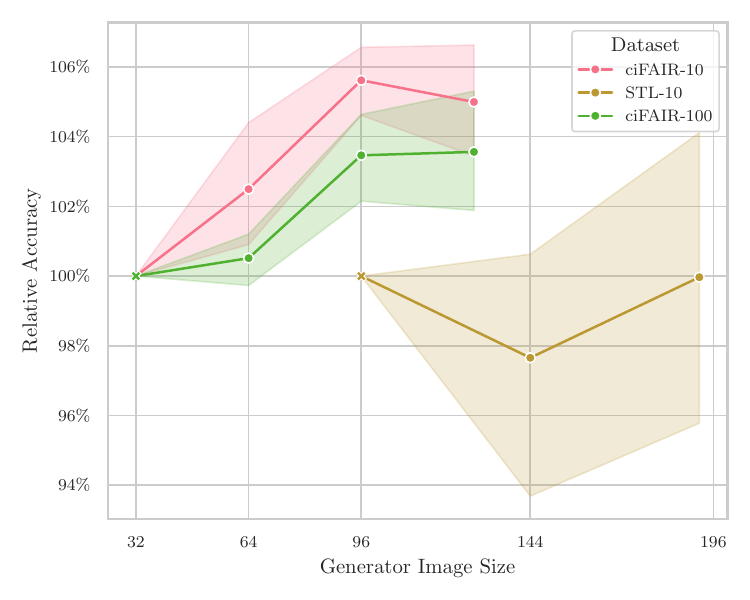}
  \caption{Analysis of the generator image size for different datasets showing the relative accuracy compared to the performance with the original image size (indicated by cross). 
  On datasets with a small image size, such as \cifairX and \cifairC, lifting the generator to a large image size achieves a performance gain. Experiments are performed with $5$ examples per class. The downstream classifier is always trained on the original image size.}
  \label{fig_ablation_image_size}
\end{figure}

\begin{figure}[!t]
    \centering
    \includegraphics[width=0.99\linewidth]{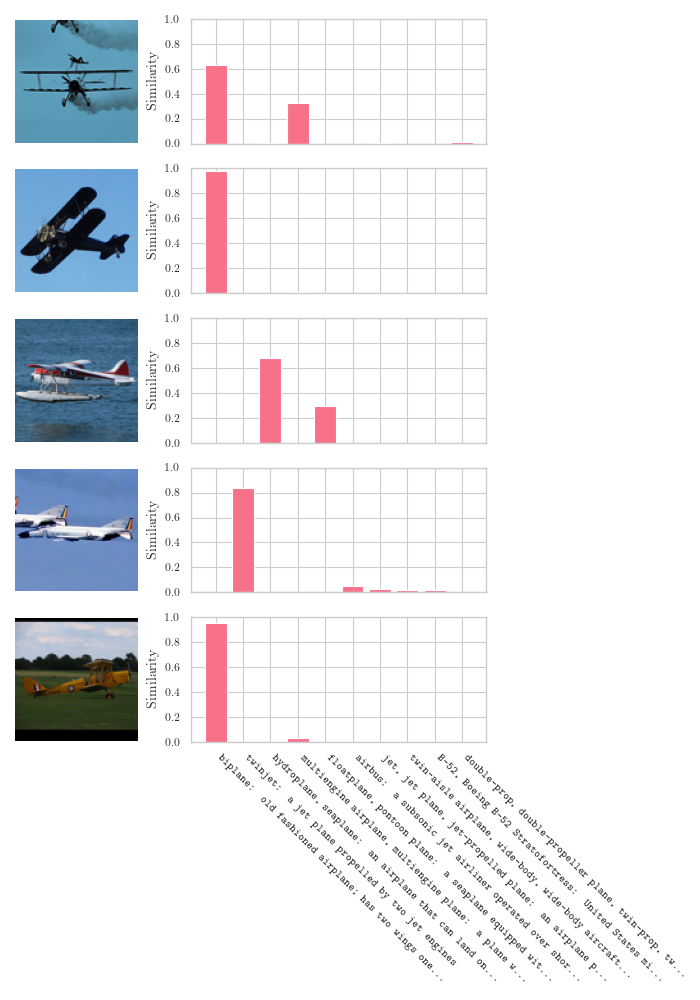}
    \caption{Distribution of the synset similarity with respect to each image. Images from the class \texttt{Airplane} of the \stl dataset and the top 10 synsets sorted by average similarity are shown. The synsets are represented by the format "$\textit{Synonym}_1, \dots, \textit{Synonym}_K\!: \textit{Description}$". The similarities are computed using the CLIP embeddings of the synset representations and the images.}
    \label{fig_clip_examples}
\end{figure}

\subsubsection{Generator Image Size}
\label{sec_analysis_generator_image_size}

In the next analysis, we investigate operating the generator on a larger image resolution as proposed in \cref{sec_method_training}. 
For that, the input images are upsampled by bilinear interpolation to a target image size before processing by the generator. Afterward, the output of the generator is downsampled to the original image size accordingly. In this process, no additional information is required, and the generator is enabled to more effortlessly produce high-frequency details, especially for datasets with a small image size. The classifier is always trained on the original image size.

The relative classification accuracy compared to the classification accuracy on the original image size with different generator image sizes is shown in \cref{fig_ablation_image_size}. The results show that for datasets with a small image size (\cifairX and \cifairC), operating the generator on a larger image size is very effective. A generator image size of $96 \times 96$, for example, boosts the performance by $5.61\%$ on \cifairX and $3.46\%$ on \cifairC. On \stl, which has larger images, increasing the generator image size to $1.5$x of the original image size leads to a slightly decreased performance, while increasing the generator image size by $2$x achieves the same performance. Thus, operating the generator on a larger image resolution is not necessary for datasets with an already large image size.

\subsubsection{Influence of Noisy Segmentation Masks}

\begin{figure}[t]
  \centering
  \includegraphics[width=\linewidth]{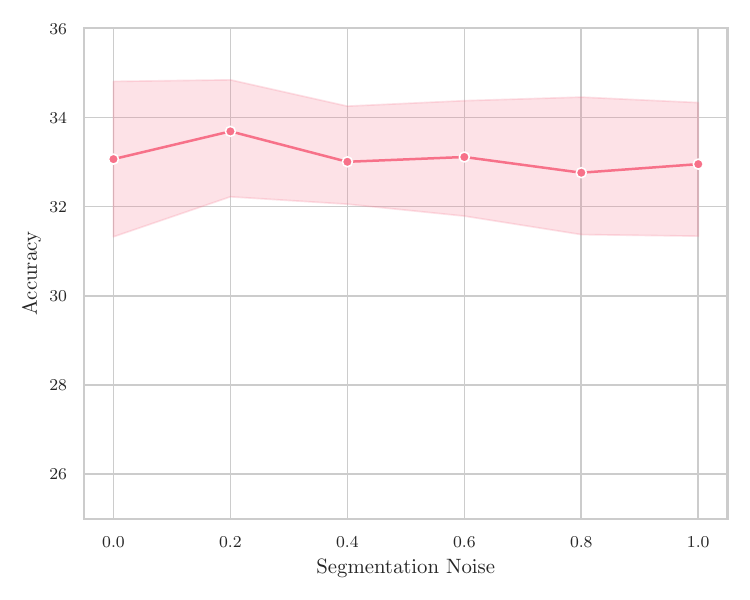}
  \caption{Analysis of the impact of noisy segmentation masks on \methodname on \stl with $5$ examples per class.}
  \label{fig_ablation_noisy_segmentation}
\end{figure}

\begin{table*}
\caption{Cross-domain analysis of \hydramix. The generator is trained on one dataset (per row) and integrated into the training of the downstream classifier on different datasets (per column).}
\label{tab:ablation_cross_domain}
\centering
\begin{tabular}{ll|S[table-format=2.2, table-space-text-post=\std{9.99}]|S[table-format=2.2, table-space-text-post=\std{9.99}]|S[table-format=2.2, table-space-text-post=\std{9.99}]|S[table-format=2.2, table-space-text-post=\std{9.99}]}
\toprule
& & \multicolumn{3}{c|}{Downstream Classifier Dataset} & {}\\
&  & {\cifairX} & {\stl} & {\cifairC} & {Average} \\
\midrule
\multirow[c]{3}{*}{\parbox{1.7cm}{Generator\\Dataset}} 
& \cifairX & 39.05\std{2.77}{} & 34.20\std{1.80} & 21.81\std{1.53} & 31.69\std{2.03}\\
& \stl & 36.42\std{3.11}  &  33.09\std{1.59}{} & 20.93\std{1.27} & 30.15\std{1.99}\\
& \cifairC & 37.39\std{3.02} & 35.72\std{1.73} &  24.86\std{0.54}{} & 32.66\std{1.76} \\
\bottomrule
\end{tabular}

\end{table*}

\begin{figure*}[ht!]
    \centering
    \includegraphics[width=0.99\linewidth]{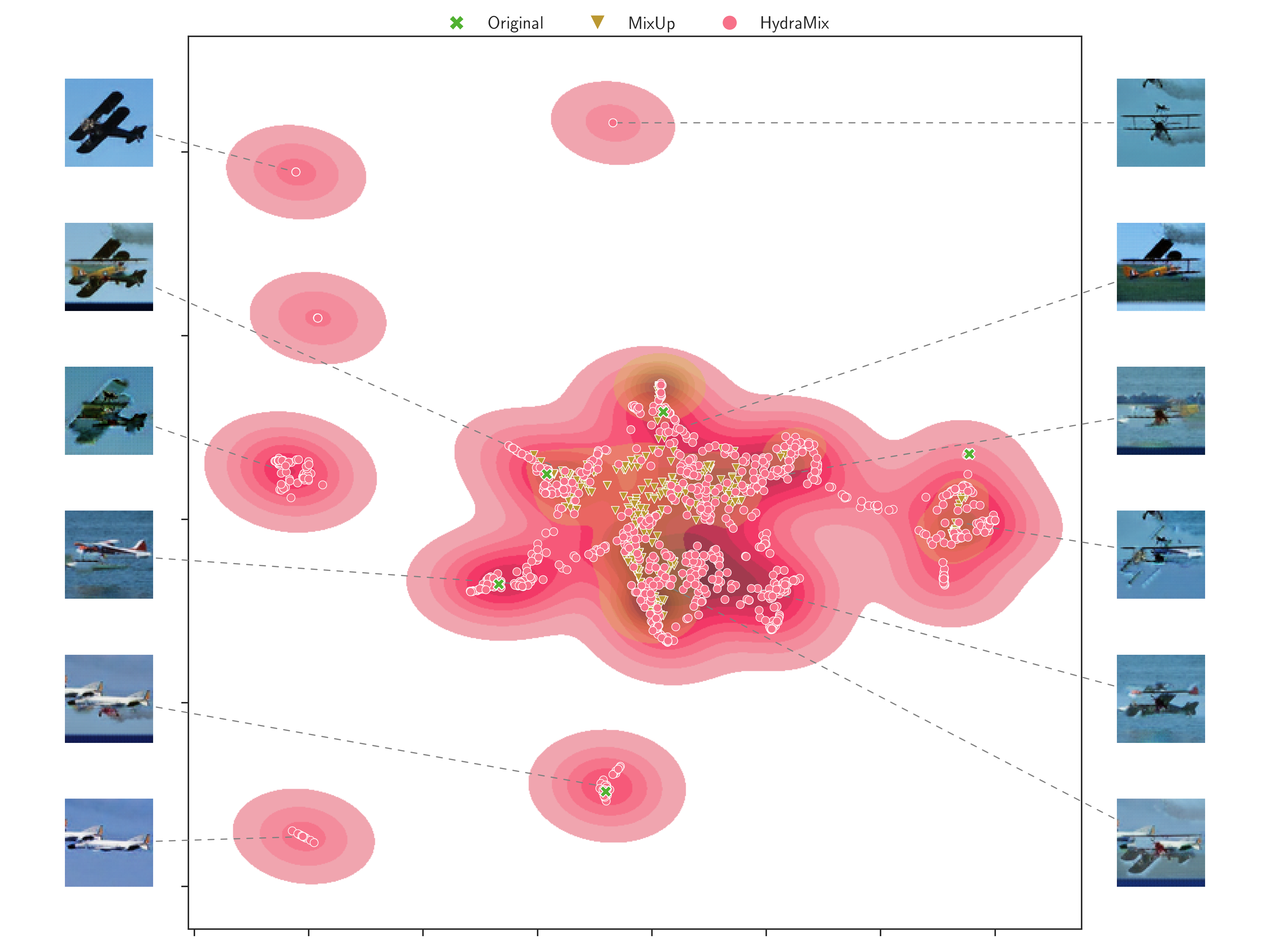}
    \caption{Visualization of the image embeddings with CLIP. The original images (green cross), MixUp images (yellow triangles), and \methodname images (red dots). The density distribution of MixUp is highlighted in yellow, and the density distribution of HydraMix is highlighted in red.}
    \label{fig_clip_features}
\end{figure*}

In the following analysis, we evaluate the influence of noisy segmentation masks. 
\chimeramixseg and \methodname sample mixing masks based on the segmentation masks. 
A Felzenszwalb Segmentation algorithm is employed here, whereas any segmentation algorithm can be integrated.
To analyze the impact of noisy segmentation masks, the masks are synthetically distorted by introducing a segmentation noise factor. With this probability, each segment is replaced with the segmentation from a different image, which is randomly selected.
Thus, the segmentation noise factor serves as a parameter, varying from the original segmentation (Segmentation Noise = 0) to a completely incorrect segmentation from a different image (Segmentation Noise = 1).
The results of \methodname on \stl with $5$ examples per class are shown in \cref{fig_ablation_noisy_segmentation}. It can be seen that \methodname is very robust to noisy segmentation masks.

\subsubsection{Cross-Domain Analysis}

In a final analysis, the cross-domain generalization of \hydramix is evaluated by training a generator on one domain and performing the downstream classification on another domains. 
The results are shown in \cref{tab:ablation_cross_domain}.
In each row, a generator is trained on the respective dataset.
Subsequently, the generators are integrated into the training of the downstream classifier across different datasets, as shown in the respective columns.
Interestingly, on \stl, the performance improves from $33.09\%$ to $34.20\%$ and $35.72\%$ when integrated the generator trained on \cifairX and \cifairC, respectively. While \cifairX has the same number of training samples, \cifairC has ten times as many training samples due to its larger number of classes.
On \cifairX and \cifairC, the best results is achieved when the generator is trained on the same dataset.
Overall, the results demonstrate that \hydramix can successfully generalize across different data distributions while maintaining competitive performance.
In the last column, the average downstream classifier accuracy across different datasets for each generator is shown.

\subsection{CLIP Features}
\label{sec_clip_feature_appendix}

To assess the generalization capabilities of the mixed generated images, we introduce the novel CLIP Synset Entropy metric $\operatorname{CSE}$ in \cref{sec_clip_synset_entropy}.
The metric is computed as the entropy of the average similarity distribution between a set of images and all hyponyms of the synset of the class. The textual representations of a synset  consist of its synonyms as well as the description.
In \cref{fig_clip_examples}, we show an example of the similarity distributions between a set of original images of the class \texttt{Airplane} and the synset representations.
It is evident that by measuring the similarity to the hyponyms of each class, we can identify sub-concepts within a class, such as \texttt{biplane}, \texttt{twinjet}, or \texttt{hydroplane}.
Therefore, the CLIP Synset Entropy serves as an effective metric for assessing the diversity within a generated dataset by calculating the coverage of the hyponyms.

To further highlight the structure of the generated images by \methodname, we visualize the CLIP embedding space of the original subsampled dataset, the embeddings of MixUp images, and the \methodname combinations using PaCMAP \citep{wangUnderstandingHowDimension2021}. An example of the class \texttt{Airplane} on \stl with $5$ samples per class is shown in \cref{fig_clip_features}. The original images are marked with a green cross. For MixUp and \methodname, the image embeddings as well as the density distributions are shown in yellow and red, respectively.
The embedding demonstrates that \methodname is able to cover a much larger portion of the space between the original samples compared to MixUp.

\subsection{Hyperparameters of \methodname}
\label{sec:hyperparameters}

\begin{table}[t]
    \centering
    \caption{Common hyperparameters of the \methodname generator for all datasets.}
    \begin{tabular}{c|c}
        Epochs & 200 \\
        Optimizer & Adam, $\beta$=[0.5, 0.999] \\
        Learning Rate Schedule & Stepwise Decay of  \\
         & \num{0.2}@[60,120,160] \\
        Initial Learning Rate & \num{0.0002} \\
        Weight Decay & \num{0.0005} \\
        Residual Blocks & 4 \\
        Feature Mixing after Block & 2 \\
        Mixing Mask Size & 4
    \end{tabular}
    \label{tab:appendix_hps_generator_common}
\end{table}

The loss terms in \cref{eq:loss} are defined as $\alpha_{\text{rec}}$ = 1000, $\alpha_{\text{per}} = 1$, and $\alpha_{\text{G,disc}} = 1$ to balance the small magnitude of the mean squared error reconstruction loss.
In \cref{sec_clip_synset_entropy}, we set the softmax temperature parameter $\tau$ to $1 / 100$ to sharpen the similarity distributions. Further hyperparameters of \methodname are shown in \cref{tab:appendix_hps_generator_common}.

\section{Conclusion}

In this work, we presented a novel method for generating new image compositions from only a handful of images that achieves state-of-the-art results in small data image classification. The architecture introduces a feature-mixing architecture that combines the content from an arbitrary number of images guided by a segmentation-based mixing mask.
Extensive experiments demonstrated the superior performance of \methodname compared to existing approaches in the small data regime. Additionally, \methodname can be successfully combined with automated augmentation methods. Finally, we presented a CLIP Synset Entropy to analyze the distribution of the generated images and showed that \methodname is able to generate a larger variety of synset concepts.
In the future, we would like to explore unsupervised segmentation methods like \citep{engelckeGENESISV2InferringUnordered2021} that are trained end-to-end with the generator to cluster the latent features and adjust them for the small data regime.

\section*{Acknowledgements}

This work was supported by the Federal Ministry of Education and Research (BMBF), Germany under the AI service center KISSKI (grant no. 01IS22093C) and the Deutsche Forschungsgemeinschaft (DFG) under Germany’s Excellence Strategy within the Cluster of Excellence PhoenixD (EXC 2122).

\bibliography{bibliography}
\bibliographystyle{icml2025}

\end{document}